\documentclass[pdflatex,sn-mathphys-num]{sn-jnl}

\usepackage{multirow,booktabs,array}
\usepackage{graphicx}%
\usepackage{multirow}%
\usepackage{amsmath,amssymb,amsfonts}%
\usepackage{amsthm}%
\usepackage{mathrsfs}%
\usepackage{textcomp}%
\usepackage{manyfoot}%
\usepackage{booktabs}%
\usepackage{algorithm}%
\usepackage{algorithmicx}%
\usepackage{algpseudocode}%
\usepackage{listings}%
\usepackage{multirow}
\usepackage{multirow}
\usepackage{graphicx} 
\usepackage{graphicx}
\usepackage{booktabs} 
\usepackage{multirow}
\usepackage{makecell}
\usepackage{pgf}
\usepackage{booktabs}
\usepackage[pagewise]{lineno}
\usepackage{pdfpages}
\theoremstyle{thmstyleone}%
\theoremstyle{thmstyletwo}%

\theoremstyle{thmstylethree}%

\begin{document}

\title[DepressLLM]{DepressLLM: Interpretable domain-adapted language model for depression detection from real-world narratives}


\author[1]{\fnm{Sehwan} \sur{Moon}}\email{sehwanmoon@etri.re.kr}
\author[1]{\fnm{Aram} \sur{Lee}}
\author[1]{\fnm{Jeong Eun} \sur{Kim}}

\author[2]{\fnm{Hee-Ju} \sur{Kang}}
\author[2]{\fnm{Il-Seon} \sur{Shin}}
\author*[2]{\fnm{Sung-Wan} \sur{Kim}}\email{shalompsy@hanmail.net}
\author*[2]{\fnm{Jae-Min} \sur{Kim}}\email{jmkim@chonnam.ac.kr}
\author*[2]{\fnm{Min} \sur{Jhon}}\email{minjhon@chonnam.ac.kr}
\author*[2]{\fnm{Ju-Wan} \sur{Kim}}\email{tarot383@naver.com}

\affil[1]{\orgdiv{AI Convergence Research Section}, \orgname{Electronics and Telecommunications Research Institute}, \orgaddress{\country{Republic of Korea}}}

\affil[2]{\orgdiv{Department of Psychiatry}, \orgname{Chonnam National University Medical School}, \orgaddress{\street{160 Baekseoro, 12 Dong-gu}, \city{Gwangju}, \postcode{61469},  \country{Republic of Korea}}}


\abstract{Advances in large language models (LLMs) have enabled a wide range of applications. However, depression prediction is hindered by the lack of large‑scale, high‑quality, and rigorously annotated datasets. This study introduces DepressLLM, trained and evaluated on a novel corpus of 3,699 autobiographical narratives reflecting both happiness and distress. DepressLLM provides interpretable depression predictions and, via its Score‑guided Token Probability Summation (SToPS) module, delivers both improved classification performance and reliable confidence estimates, achieving an AUC of 0.789, which rises to 0.904 on samples with confidence $\geq$ 0.95. To validate its robustness to heterogeneous data, we evaluated DepressLLM on in‑house datasets, including an Ecological Momentary Assessment (EMA) corpus of daily stress and mood recordings, and on public clinical interview data. Finally, a psychiatric review of high‑confidence misclassifications highlighted key model and data limitations that suggest directions for future refinements. These findings demonstrate that interpretable AI can enable earlier diagnosis of depression and underscore the promise of medical AI in psychiatry.}




\maketitle

\section{Introduction}\label{sec1}

Depression, a highly prevalent mental disorder, is projected to become a leading contributor to the global disease burden by 2030 \cite{malhi2018course}.
Because language use reflects emotional states, language-based approaches for depression-screening tools are increasingly regarded as noninvasive and cost-effective alternatives.
Numerous studies have examined the language patterns of individuals with depression, demonstrating a strong association between language use and depression \cite{behdarvandirad2022depression,hur2024language,weintraub2023word, kazmierczak2024natural,hartnagel2025linguistic}. 

With recent advancements in artificial intelligence (AI), large language models (LLMs) have demonstrated remarkable capabilities in a wide range of natural language processing tasks.
Trained on massive datasets of text and code, LLMs can perform diverse functions, such as language translation \cite{bahdanau2014neural}, text summarization \cite{ouyang2022training}, question answering \cite{bogireddy2024comparative}, and code generation \cite{li2022competition}.

However, current research on depression screening using LLMs is limited owing to the lack of clinically validated diagnostic datasets. Researchers have used diverse textual modalities for depression detection, ranging from personal diary entries \cite{shin2024using} to clinical interview transcripts combined with facial feature analysis \cite{sadeghi2024harnessing}, and posts from Reddit and other social media to mine linguistic markers of depression \cite{wang2024explainable}.
Transformer-based models, such as MentalBERT \cite{ji2021mentalbert} and MentalLLaMA \cite{yang2024mentallama}, have been fine-tuned on large-scale, social media-derived mental health corpora to effectively detect depression. However, these studies often rely on datasets with human assessments inferred from social media rather than standardized annotations such as the Patient Health Questionnaire-9 (PHQ-9) \cite{kroenke2001phq} or the Beck Depression Inventory \cite{beck1996beck}. This often leads to a reliance on labels derived from specific keywords or self‑reported diagnostic statements \citep{coppersmith2014quantifying} and the problematic assumption of unlabeled users as healthy controls \citep{chancellor2020methods}, both of which can introduce substantial noise.

\begin{figure*}[h]
  \centering
  \includegraphics[width=\textwidth,angle=0]{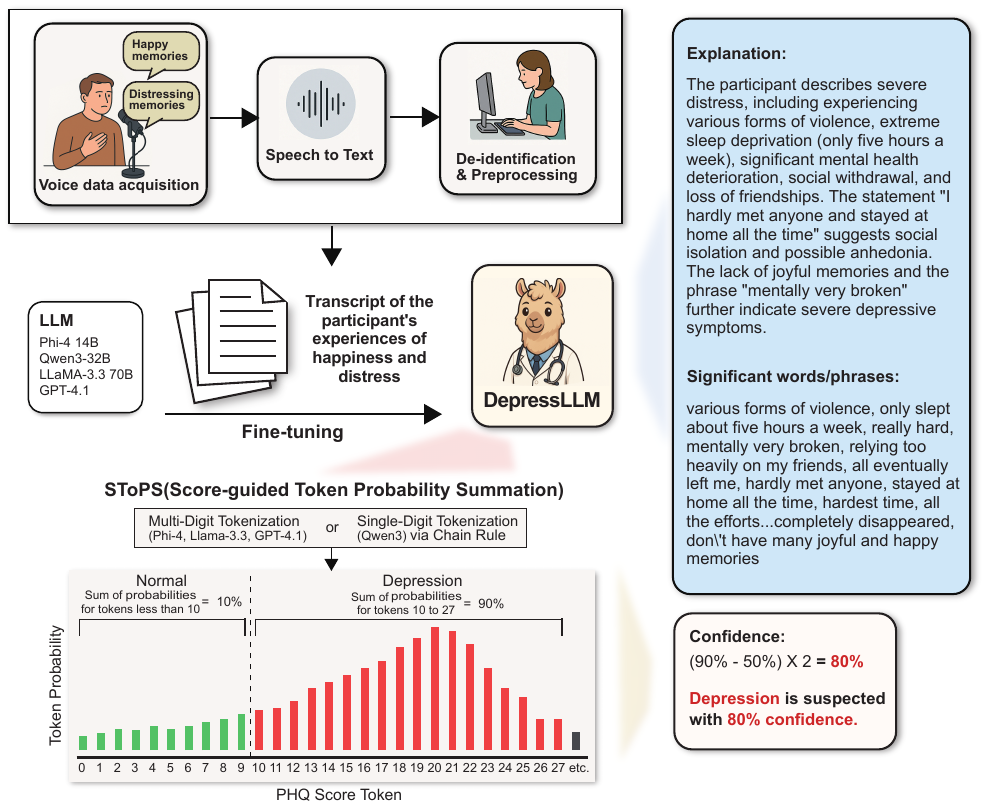} 
  \caption{\textbf{Illustration of DepressLLM.} DepressLLM is a depression-detection system based on domain-adapted LLMs. Participants provide audio recordings describing both happy and distressing memories. These recordings undergo transcription, de-identification, and preprocessing. Subsequently, this curated corpus of transcripts is used to fine-tune foundation models, culminating in the development of DepressLLM. During inference, the model leverages Score-guided Token Probability Summation (SToPS) to produce a probability distribution across PHQ-9 score tokens (0–27). The cumulative probability mass less than the clinical cutoff indicates a “normal” status, whereas the mass greater than that indicates “depression”. The disparity between these values provides a confidence score. Furthermore, DepressLLM generates concise natural-language explanations and identifies the significant phrases that informed its judgment.}\label{fig1}
\end{figure*}

We introduce DepressLLM (Figure~\ref{fig1}), a depression-detection framework trained on patient narratives to capture the linguistic hallmarks of depression. The model was trained on TREND-P, a large-scale dataset comprising 3,699 transcribed audio recordings of autobiographical memories, and evaluated on two independent datasets: VEMOD, a 265-sample Ecological Momentary Assessment (EMA) corpus, and the public Distress Analysis Interview Corpus with Wizard-of-Oz (DAIC-WOZ) \cite{gratch2014distress}. TREND-P and VEMOD are in-house datasets collected from real-world clinical and observational studies. Our model demonstrated strong and consistent classification performance across all datasets. By generating reliable confidence scores and concise natural‑language explanations alongside each prediction, DepressLLM enhances transparency and trust, representing a substantial advancement in automated mental health assessments.  DepressLLM could  enhance individuals’ self-awareness of their mental health and support earlier screening in both clinical and community settings.

\begin{table}[h]
\label{tab:char}
\caption{\textbf{Characteristics of the in-house  (TREND-P, VEMOD) and public (DAIC-WOZ) datasets} }\label{tab1}
\begin{tabular*}{\textwidth}{@{\extracolsep\fill}llll}
\toprule
Characteristics                                  & TREND-P      & VEMOD        & DAIC-WOZ         \\ \midrule
Number of participants, n                      &          3,699     &          265       &        189        \\\addlinespace[2.5pt]
Sex, n (\%)                      &               &                 &                \\
\quad Male                                             & 1,238 (33.5\%) & 62 (23.4\%)     & 102 (54.0\%)   \\
\quad Female                                           & 2,461 (66.5\%) & 203 (76.6\%)    & 87 (46.0\%)    \\\addlinespace[2.5pt]
Age group (years), n (\%)  &                 &                \\
\quad 20–39                                            & 722 (19.5\%)  & 65 (24.5\%)     & -              \\
\quad 40-59                                            & 784 (21.2\%)  & 198 (74.7\%)    & -              \\
\quad 60+                                              & 2,168 (58.6\%) & 2 (0.8\%)       & -              \\ \addlinespace[2.5pt]
PHQ score, mean (SD) & 4.6 (5.3)     & 5.4 (5.1)       & 6.7 (5.9)      \\  \addlinespace[3pt]
\quad PHQ: 0–4, n (\%)                                  & 2,296 (62.1\%) & 137 (51.7\%)    & 86 (45.5\%)    \\ 
\quad PHQ: 5–27, n (\%)                                 & 1,403 (37.9\%) & 128 (48.3\%)    & 103 (54.5\%)   \\ \addlinespace[2.5pt]
\quad PHQ: 0–9, n (\%)                                  & 3,137 (84.8\%) & 219 (82.6\%)    & 132 (69.8\%)   \\
\quad PHQ: 10–27, n (\%)                                & 562 (15.2\%)  & 46 (17.4\%)     & 57 (30.2\%)    \\ \addlinespace[2.5pt]
Number of tokens, mean (SD)                              & 364.3 (174.3) & 3,143.4 (1260.8) & 2,756.3 (999.2) \\ \botrule
\end{tabular*}
\footnotesize SD = standard deviation; DAIC-WOZ uses PHQ-8 (score range 0–24).
\end{table}

\section{Results}\label{sec2}
\subsection{Overall Design of DepressLLM}

Figure~\ref{fig1} illustrates the overall design of DepressLLM, a system based on domain-adapted language models. We utilized the TREND-P dataset, a multimodal dataset  collected by the authors, comprising psychological scales, interview videos with audio, heart rate variability, and vital signs, including blood and actigraphy data.  From this dataset, we extracted and transcribed 3,699 audio recordings in which participants recounted happy and distressing memories, producing a balanced corpus of first-person narratives featuring paired positive and negative contexts and fine-grained emotional language. In-house audio recordings were transcribed, de-identified, and preprocessed, culminating in a curated corpus of 3,699 transcripts, 80\% of which were used to fine-tune the models. 

To develop DepressLLM, we utilized both proprietary and open-source foundation models. We fine-tuned OpenAI’s models \cite{brown2020language, achiam2023gpt} to explore their performance with a high-capacity proprietary model and simultaneously trained open-source variants to ensure reproducibility and public accessibility. This approach enables robust performance benchmarking against state-of-the-art LLMs while providing a shareable version suitable for open research and deployment. 

During inference, DepressLLM receives narrative input and predicts a probability distribution over the PHQ-9 score tokens (0–27).
Our proposed Score-guided Token Probability Summation (SToPS) method then generates a binary classification and a confidence score.
A cumulative probability mass less than the clinical cutoff signifies a “normal” status, whereas a mass greater than that indicates “depression”. The disparity between these values provides an intuitive confidence score. Furthermore, DepressLLM generates concise natural-language explanations that clearly articulate the reasoning behind its judgment and identifies the significant phrases that informed its decision, thereby offering insight into its reasoning process. 

For evaluation, we assessed DepressLLM on three datasets. First, a held-out portion (20\%) of the TREND-P dataset, distinct from the training set, was used to evaluate the in-domain classification performance. Second, we tested the model using VEMOD, an internally collected Ecological Momentary Assessment dataset comprising 265 transcribed daily voice recordings describing participants’ momentary stress and mood states. Third, we evaluated the model using the public DAIC-WOZ corpus, a benchmark clinical interview dataset commonly used in affective computing and mental health research.  Table~\ref{tab1} summarizes the key characteristics of the datasets. As listed in the table, these datasets differ not only in task paradigm but also in sample demographics, clinical characteristics, and token lengths, enabling us to assess the robustness of DepressLLM across heterogeneous linguistic and contextual settings. 

The subsequent sections of this report detail our findings on (i) classification performance on the TREND-P dataset, (ii) classification performance across heterogeneous datasets, (iii) classification performance based on confidence thresholds, (iv) lexical evidence underlying the predictions, and (v) analysis of high-confidence errors with psychiatric validation.

\subsection{Evaluation of classification performance}

We evaluated depression prediction performance by considering two factors: training strategy (zero-shot prompting, supervised learning, and fine-tuning) and classification type (score-based, binary, and SToPS-based). In the score-based approach, each model first generates a PHQ-9 score (ranging from 0 to 27). Depression is then determined using a clinical PHQ score cutoff of 5 or 10. In the binary approach, the model classifies the outcome as “normal” or “depression”. For SToPS-based classification, the model outputs a calibrated probability mass over PHQ score tokens, where the cumulative probability less than a clinical cutoff indicates “normal” and the remaining mass indicates “depression,” enabling alignment with the clinical cutoff. We report performance using the threshold-independent area under the receiver operating characteristic curve (AUC), which reflects overall discriminative ability, and the threshold-dependent, class-imbalance-robust Matthews correlation coefficient (MCC) \cite{chicco2020advantages}. 

 We evaluated the baseline depression classification performance of widely adopted LLMs in a zero-shot setting. At a clinical cutoff of 10, GPT-4.5 achieved the highest overall classification performance (AUC = 0.749, MCC = 0.310). At a clinical cutoff of 5, GPT-4.5 attained the highest AUC of 0.716, whereas o1-pro \cite{jaech2024openai} produced the most balanced classification result, achieving the highest MCC of 0.367. All GPT-4 family models outperformed the GPT-3.5-turbo model \cite{brown2020language} across all settings. Among the open-source models, LLaMA-3.3 70B~\cite{grattafiori2024llama} demonstrated superior performance compared to both Microsoft’s Phi-4 27B~\cite{abdin2024phi} and Alibaba's Qwen3-32B \cite{yang2025qwen3} and outperformed GPT-3.5-turbo, highlighting its competitiveness despite being an open-source model. Classification performance for distressing memories was consistently higher than for happy memories, and using both types of memories together yielded the strongest results. Because the binary approach directly partitions outputs into classes without relying on a clinical cutoff, we compared performance solely by AUC and found that the score‑based approach yielded higher AUC values. Supervised binary classification using sentence embeddings and machine learning yielded performance comparable to zero-shot prompting. Previous domain-adapted language models for mental health classification, such as MentalBERT and MentalRoBERTa \cite{ji2021mentalbert}, outperformed GPT-3.5-turbo, Phi-4, and Qwen3 when evaluated in a zero-shot prompting setting. MentaLLaMA-chat-13B, an instruction-tuned model from \cite{yang2024mentallama}, exhibited limited performance, often ignoring task prompts and failing to produce clear classifications in 303 of the 740 instances. Our model achieved state-of-the-art results across all evaluation settings. Performance varied with the backbone architecture; fine-tuning GPT-4 resulted in substantial improvements over GPT-3.5. Both LLaMA-3.3 and Phi-4 outperformed GPT-3.5-turbo as backbone models for DepressLLM. LLaMA-3.3, which has a significantly larger number of parameters than Phi-4, demonstrated higher classification performance.
For our DepressLLM model, incorporating explanatory rationales and self-reported confidence scores into the output had a minimal impact on the overall classification performance.
However, removing the SToPS module resulted in a drop of 0.050 in AUC and 0.075 in MCC.
The prompt instructions employed in this experiment are provided in Supplementary section 1.

\begin{table}[t]
\tiny
\setlength{\tabcolsep}{1pt} 
\caption{\textbf{Classification performance across models and training strategies at clinical PHQ-9 cutoffs of 5 and 10.}}
\label{tab2}
\begin{tabular*}{\textwidth}{@{\extracolsep\fill}lllcccc}
\toprule
 \multirow[c]{2}{*}{\begin{tabular}[c]{@{}c@{}}Training\\ strategy\end{tabular}}
  & \multirow[c]{2}{*}{\begin{tabular}[c]{@{}c@{}}Classification\\ type\end{tabular}}
  & \multirow[c]{2}{*}{Model}
  & \multicolumn{2}{c}{Cutoff=5}
  & \multicolumn{2}{c}{Cutoff=10} \\
\cmidrule(r){4-5}\cmidrule(r){6-7}
  & & 
  & AUC & MCC & AUC & MCC \\
\midrule
\multirow{14}{*}{Zero-shot}                                                    & \multirow{12}{*}{\begin{tabular}[c]{@{}c@{}}Score-based\\ classification\end{tabular}}
                                                                                & Phi-4 14B                                 & 0.652         & 0.228        & 0.646          & 0.175         \\& & Qwen3-32B                                  & 0.688         & 0.215         & 0.706          & 0.157         \\  
                                                                                                                                                               & & LLaMA-3.3 70B                                 & 0.700         & 0.290         & 0.739          & 0.275         \\  

&    & GPT-3.5 turbo                              & 0.637         & 0.221         & 0.663          & 0.210         \\  
&                                                                                        & GPT-4o (Happy)                             & 0.573         & 0.186         & 0.588          & 0.082         \\
                                                                               &                                                                                        & GPT-4o (Distress)                          & 0.655         & 0.117         & 0.662          & 0.124         \\
                                                                               &                                                                                        & GPT-4o                                     & 0.701         & 0.165         & 0.730          & 0.226         \\                                                      
                                                                               &                                                                                        & GPT-4.5                                    & 0.716         & 0.273         & 0.749          & 0.310         \\ &                                                                                        & GPT-4.1 (Happy)                            & 0.574         & 0.149         & 0.578          & 0.115         \\
                                                                               &                                                                                        & GPT-4.1 (Distress)                         & 0.668         & 0.257         & 0.691          & 0.219         \\
                                                                               &                                                                                        & GPT-4.1                                    & 0.709         & 0.320         & 0.724          & 0.265         \\ 
                                                                               &                                                                                        & o3-mini                                    & 0.678         & 0.292         & 0.700          & 0.244         \\
                                                                               &                                                                                        & o1-pro                                     & 0.699         & 0.367         & 0.731          & 0.259         \\
                                                                       \cmidrule(r){2-2}

                                                                              &                                                                                        & GPT-4o (Binary classification)                                & 0.696        & 0.209         & 0.727          & 0.178  \\       &                                                                                        & GPT-4.1 (Binary classification)                                & 0.648        & 0.279         & 0.692          & 0.268         
\\ \cmidrule(r){1-1} \multirow{3}{*}{\begin{tabular}[c]{@{}c@{}}Supervised\\ learning\end{tabular}} & \multirow{5}{*}{\begin{tabular}[c]{@{}c@{}}Binary\\ classification\end{tabular}}                                                & Embedding (Happy) + ML                     & 0.644$\pm$0.010&         0.229$\pm$0.008       & 0.742$\pm$0.015     & 0.211$\pm$0.036        \\
                                                                               &                                                                                        & Embedding (Distress) + ML                  & 0.679$\pm$0.014         & 0.264$\pm$0.018         & 0.726$\pm$0.015          & 0.255$\pm$0.050        \\
                                                                               &                                                                                        & Embedding + ML                             & 0.688$\pm$0.004          & 0.279$\pm$0.008        & 0.762$\pm$0.003           & 0.218$\pm$0.019         \\ \cmidrule(r){1-1}
\multirow{11}{*}{Fine-tuning}                                                  &                                                                                        & MentalBERT                                 & 0.686$\pm$0.016&0.273$\pm$0.021&0.735$\pm$0.012&0.220$\pm$0.031
         \\
                                                                               &                                                                                        & MentalRoBERTa                              & 0.693$\pm$0.004&0.256$\pm$0.007&0.748$\pm$0.008&0.255$\pm$0.022
        \\
                                                                               &                                                                                        & MentaLLaMA-chat-13B\textsuperscript{\ddag}                         & -         & 0.179
         & -         & 0.075
         \\
                                                                               &                                                                                        & \makecell[l]{DepressLLM  \\(GPT-4.1, w/o SToPS)}                &0.774$\pm$0.006&0.405$\pm$0.021&0.807$\pm$0.015&0.374$\pm$0.016

      \\  \cmidrule(r){2-2}
                                                                               & \multirow{7}{*}{\begin{tabular}[c]{@{}c@{}}SToPS-based\\ classification\end{tabular}}  & DepressLLM (Phi-4 14B)                  & 0.775$\pm$0.008&0.414$\pm$0.010&0.828$\pm$0.004&0.363$\pm$0.018
        \\ 
                                                                               &                                                        & DepressLLM (Qwen3-32B)                         &0.776$\pm$0.003&0.398$\pm$0.007&0.815$\pm$0.004&0.298$\pm$0.034
         \\     
                                                                               &                                                                                                    & DepressLLM (LLaMA-3.3 70B)                         &0.779$\pm$0.007&0.420$\pm$0.009&0.830$\pm$0.007&0.391$\pm$0.029
         \\     
                                                                               &                                                                                        & DepressLLM (GPT-3.5 turbo)                      & 0.767$\pm$0.001&0.385$\pm$0.009&0.810$\pm$0.007&0.368$\pm$0.015
         \\
                                                                               &                                                                                        & DepressLLM (GPT-4o)                        & 0.787$\pm$0.001&0.415$\pm$0.009&0.846$\pm$0.006&0.431$\pm$0.033 \\
                                    &    & DepressLLM (GPT-4.1)                        &  0.789$\pm$0.003&0.425$\pm$0.022&0.835$\pm$0.004&0.415$\pm$0.034 
        \\
                                                                               &                                                                                        &+Explanation                                     & 0.786$\pm$0.005&0.414$\pm$0.015&0.834$\pm$0.003&0.399$\pm$0.013

         \\
                                                                               &                                                                                        &+Explanation+ Self‑reported Conf.       & 0.788$\pm$0.003&0.430$\pm$0.013&0.833$\pm$0.002&0.400$\pm$0.002
    \\
\botrule
\end{tabular*}
\footnotesize{\textsuperscript{\ddag}Excluded 303 instances where prompts were ignored or no classification was returned.}
\end{table}

\subsection{Generalization to Heterogeneous Datasets}

To evaluate the robustness of DepressLLM across heterogeneous data types, we tested the model on two datasets: (1) VEMOD narratives related to stress and mood, which were independently collected for this study, and (2) clinical interviews from the public DAIC-WOZ corpus \cite{gratch2014distress}. As listed in Table~\ref{tab3}, DepressLLM consistently outperformed the GPT-4.1 baseline across both the VEMOD and DAIC-WOZ datasets at clinical cutoffs of 5 and 10. On the VEMOD dataset, DepressLLM outperformed GPT-4.1 by 0.034 in AUC and 0.044 in MCC, and by 0.046 in AUC and 0.076 in MCC at clinical cutoffs 5 and 10, respectively. On the DAIC-WOZ dataset, DepressLLM achieved an AUC of 0.920 and an MCC of 0.619 at a clinical cutoff of 5, outperforming the GPT-4.1 baseline by a notable margin. This advantage was maintained at a clinical cutoff of 10, demonstrating the reliability of the model on external data. Incorporating the SToPS method improved model performance across all evaluation settings, except for the VEMOD dataset at a clinical cutoff of 10, where a slight decrease in AUC was observed.

\begin{table}[t]
\caption{\textbf{Performance comparison on heterogeneous datasets (VEMOD and DAIC-WOZ)}}
\label{tab3}
\setlength{\tabcolsep}{1pt} 
\begin{tabular*}{\textwidth}{@{\extracolsep\fill}llcccc}
\toprule
 \multirow[c]{2}{*}{\begin{tabular}[c]{@{}c@{}}Testing\\ dataset\end{tabular}}
  & \multirow[c]{2}{*}{Model}
  & \multicolumn{2}{c}{Cutoff=5}
  & \multicolumn{2}{c}{Cutoff=10} \\
\cmidrule(r){3-4}\cmidrule(r){5-6}
  &   
  & AUC & MCC & AUC & MCC \\
\midrule
\multirow{3}{*}{VEMOD dataset} 
                         & GPT-4.1                                     & 0.732        & 0.337        & 0.784        & 0.318        \\
                                                                               &                                                     DepressLLM (w/o SToPS)                                    & 0.755$\pm$0.011        & 0.380$\pm$0.031         & 0.832$\pm$0.014        & 0.372$\pm$0.010         \\
                                                                               &                                                     DepressLLM                                     & 0.766$\pm$0.005        & 0.381$\pm$0.033         & 0.830$\pm$0.008          & 0.394$\pm$0.039         \\\addlinespace[5.5pt]

\multirow{3}{*}{\begin{tabular}[c]{@{}c@{}}DAIC-WOZ\\ dataset\end{tabular}}  & GPT-4.1                   & {0.894}        & {0.576}      & {0.875}         & {0.550}         \\                                                                                & DepressLLM (w/o SToPS)                  & 0.899$\pm$0.006       & 0.597$\pm$0.013        & 0.861$\pm$0.003         & 0.534$\pm$0.039        \\
                                                                                                                    & DepressLLM                            & 0.920$\pm$0.005        & 0.619$\pm$0.034       & 0.880$\pm$0.001         & 0.566$\pm$0.018        \\ 
\botrule
\end{tabular*}
\end{table}

\subsection{Internal and External Evaluation of Confidence Calibration with SToPS}

\begin{figure*}[h]
  \centering
  \includegraphics[width=\textwidth,angle=0]{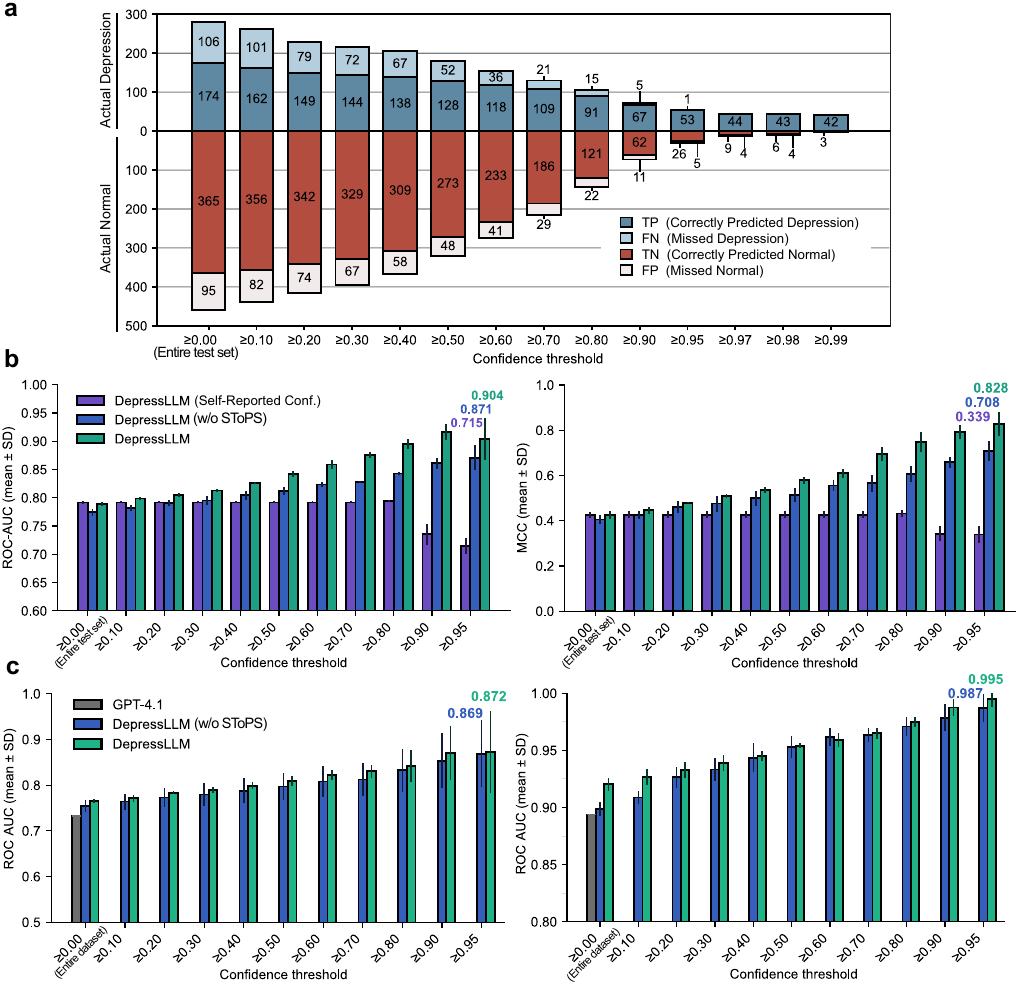} 
  \caption{
  \textbf{Confidence-based filtering using the SToPS method.}
  (a) Confusion matrix counts on the TREND-P test set ($n$ = 740) across varying confidence thresholds. 
  (b) Comparison of confidence estimation methods on TREND‑P data based on AUC (left) and MCC (right). (c) Comparison of AUC performance when evaluated on the VEMOD dataset (left) and the DAIC-WOZ dataset (right) as test sets.
  }\label{fig2}
\end{figure*}

The predictive reliability of the calibrated confidence scores generated by the SToPS method was evaluated using an internal TREND-P test set comprising 740 participants (Figure~\ref{fig2}a).
As the confidence threshold increased, uncertain predictions were progressively excluded, leading to improved classification accuracy. All cases were included at a default threshold of 0, yielding an accuracy of 72.8$\%$. Applying a confidence threshold of 0.5 retained 67.7$\%$ of the samples, which achieved 80.0$\%$ accuracy. When the threshold was increased to 0.95, 11.5$\%$ of the samples remained and showed 92.9$\%$ accuracy. These results show that higher confidence scores correspond to higher accuracy, indicating that SToPS provides a reliable confidence score.

Figure~\ref{fig2}b compares three confidence approaches: (1) self-reported confidence, obtained by prompting the model to state its own certainty; (2) binary logit-normalized probability, labeled “DepressLLM w/o SToPS”, derived by normalizing the logits of the “0” and “1” answer tokens; and (3) the proposed SToPS method, labeled “DepressLLM”, which aggregates token-level probabilities across PHQ-9 score tokens.  The performance based on self-reported confidence degrades as the confidence threshold increases, suggesting that the model's confidence estimates are unreliable. The binary logit-normalized probability approach yields more reliable confidence estimates but still remains less efficient than the SToPS method. By contrast, SToPS achieves higher AUC values at every threshold, consistently demonstrating more robust classification results.

To assess external validity, the same analysis was conducted on two heterogeneous datasets (Figure~\ref{fig2}c). In both cases, the proposed SToPS method outperformed the binary logit-normalized baseline and GPT-4.1. At a 0.95 confidence threshold, SToPS achieved an AUC of 0.872 and 0.995 on the VEMOD and DAIC-WOZ datasets, respectively. These results demonstrate that SToPS preserves predictive reliability across heterogeneous data. Comparisons with other confidence estimation approaches, including entropy-based, max probability--based, and margin-based methods, are provided in the Supplementary Figure S1.

\subsection{Lexical evidence underlying DepressLLM predictions}
\begin{figure*}[h]
  \centering
  \includegraphics[width=\textwidth,angle=0]{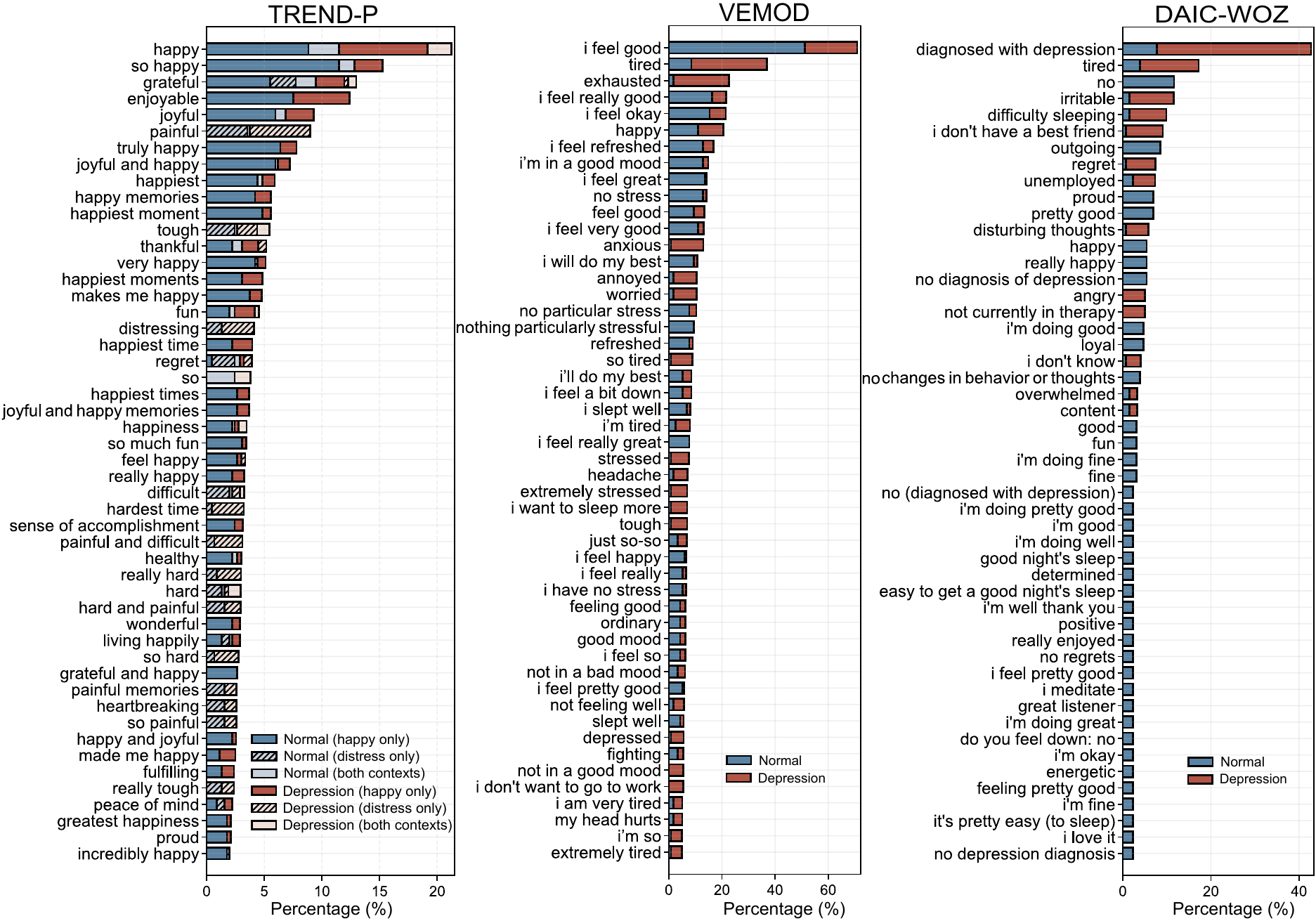} 
\caption{\textbf{Class-wise frequency comparison of DepressLLM-reported significant words/phrases across the three datasets (TREND-P, VEMOD, and DAIC-WOZ).} 
Each bar represents the percentage frequency of a word or phrase within a predicted class. For the TREND‑P dataset, we additionally stratified entries by the type of memory prompt associated with each entry (Happy‑only, Distress‑only, or Both), yielding six mutually exclusive groups for comparison.}\label{fig4}
\end{figure*}

To identify the lexical cues underlying the model's predictions, we analyzed the significant words and phrases extracted by DepressLLM for each input and summarized their class-normalized frequencies (Figure~\ref{fig4}). Compared to other datasets, the TREND-P dataset exhibited a lower tendency for specific phrases to appear exclusively in either the normal or depression prediction, likely because it includes both positive and negative narratives from each individual.  Consequently, positive (e.g., \textit{happy} and \textit{enjoyable}) and negative (e.g., \textit{painful}) expressions were prevalent across both predicted labels. Hardship-related terms such as \textit{tough} and \textit{hard}, when found in both happy and distress contexts, led to a depression prediction. In an additional analysis (see Supplementary Figure S2) focusing on significant words/phrases,  we found that in the TREND-P dataset, positive terms were more prevalent and the language used in both “normal” and “depression” predictions showed less polarization, indicating a more balanced distribution of terms across the two classes.

The VEMOD dataset, drawn from daily recordings, contains abundant mood- and stress-related expressions. Positive phrases (e.g., \textit{I feel good} and, \textit{no stress}) were dominant in the “normal” predictions, whereas negative phrases (e.g., \textit{exhausted}, \textit{anxious}, and \textit{worried}) were dominant in the “depression” predictions, resulting in a clearer lexical separation than in TREND-P.
The DAIC-WOZ dataset exhibited the clearest distinction between the predicted labels. Clinical interviews frequently contain explicit symptom-related language (e.g., \textit{no diagnosis of depression}, \textit{difficulty sleeping}, and \textit{not currently in the therapy}), which enables keywords to exert a direct influence on the model’s predictions.

\subsection{Analysis of High-Confidence Errors with Psychiatric Validation}

We conducted a detailed analysis of 16 misclassified samples among the high-confidence predictions (confidence score$\geq$0.95) generated by the DepressLLM$+$Explanation model. Two board-certified psychiatrists (M.J.; Psychiatrist A and J.W.K.; Psychiatrist B) independently reviewed these samples and assessed whether the model’s predictions aligned with their clinical judgment. Their evaluations, along with key physician comments regarding the model’s reasoning and discrepancies between model predictions and self-reported PHQ-9 scores, are summarized in Table~\ref{tab4}. Additional details can be found in Supplementary Table S1. 

Of the 16 samples, both psychiatrists determined 12 to be consistent with the model’s predictions, indicating that the clinicians agreed with the model’s interpretation rather than the self-reported PHQ-9 scores In the remaining 4 cases, the psychiatrists disagreed with each other in their assessment; however,  no cases existed in which both clinicians determined the model's prediction to be incorrect. 
Although most explanations were deemed clinically appropriate, several comments highlighted areas for improving the model’s reasoning.
Common limitations included insufficient consideration of temporal context, a lack of attention to protective or resilience factors, and challenges in distinguishing between pathological and non-pathological emotional responses. In cases with limited linguistic content, the model occasionally failed to express appropriate uncertainty, underscoring the need for explanatory mechanisms that better reflect content limitations. 

In addition, the psychiatrists  identified several potential reasons for the discrepancies between the model predictions and self-reported PHQ-9 scores. These included somatization, limited emotional awareness, and reduced insight into one’s affective state, all of which may have led participants to report their mood symptoms inaccurately while completing the PHQ-9.

\begin{sidewaystable*}[htbp]
\caption{Manual review of high-confidence misclassified cases by two clinical psychiatrists.}
\centering
\footnotesize
\begin{tabular*}{\textwidth}{@{\extracolsep{\fill}} l c c c c c p{9.5cm} }
\toprule
\makecell{Psychiatrist\\agreement} & \makecell{Case\\\#}    & \makecell{Actual\\PHQ-9}&  \makecell{Predicted\\PHQ-9} & \makecell{Psychiatrist\\A} & \makecell{Psychiatrist\\B}  & Key physician comments (condensed) \\
\midrule

\multirow{2}{*}{\begin{tabular}[c]{@{}l@{}}Agreement\\between\\psychiatrists\\(Norm)\end{tabular}} 
  & 1 & \makecell[t]{5\\(Dep)}  & \makecell[t]{0\\(Norm)}   & Normal & Normal & Appropriate model judgment. PHQ-9 likely reflects participants’ non-mood symptoms. \\
  \addlinespace[0.2pt]
  & 2 & \makecell[t]{6\\(Dep)}  & \makecell[t]{0\\(Norm)}   & Normal & Normal & Model’s judgment is appropriate based on interview content alone, but the limited mention of distress appears to make accurate assessment difficult. \\
 \addlinespace[3pt] \midrule
\addlinespace[3pt]

\multirow{24}{*}{\begin{tabular}[c]{@{}l@{}}Agreement\\between\\psychiatrists\\(Dep)\end{tabular}} 
  & 3 &  \makecell[t]{4\\(Norm)} & \makecell[t]{10\\(Dep)} & \makecell[t]{Moderate\\(Dep)}  & \makecell[t]{Severe\\(Dep)}   & Model judgment is appropriate. Direct mention of sadness and depression suggests possible misreporting on PHQ-9 items.\\   \addlinespace[0.2pt]
  & 4 & \makecell[t]{4\\(Norm)} & \makecell[t]{10\\(Dep)} & \makecell[t]{Mild\\(Dep)}     & \makecell[t]{Moderate\\(Dep)}  & Appropriate model judgment, but past events are overemphasized. Low PHQ-9 likely due to participants’ poor mood awareness.\\   \addlinespace[0.2pt]
  & 5 & \makecell[t]{1\\(Norm)}  & \makecell[t]{13\\(Dep)} & \makecell[t]{Mild\\(Dep)}     & \makecell[t]{Mild\\(Dep)}     & Content is insufficient to assess mood, but model judgment is appropriate given childhood adversity. The model should have noted the lack of information.\\   \addlinespace[0.2pt]
  & 6 &  \makecell[t]{4\\(Norm)} & \makecell[t]{8\\(Dep)} & \makecell[t]{Mild\\(Dep)}     & \makecell[t]{Mild\\(Dep)}     & Model judgment is appropriate. Reason for participants’ low PHQ-9 score is unclear.\\   \addlinespace[0.2pt]
  & 7 & \makecell[t]{1\\(Norm)}  & \makecell[t]{10\\(Dep)} & \makecell[t]{Mild\\(Dep)}       & \makecell[t]{Moderate\\(Dep)}  & Model’s judgment is appropriate, but key words should account for temporal context. Low PHQ-9 may reflect somatization.\\   \addlinespace[0.2pt]
  & 8 & \makecell[t]{2\\(Norm)}  & \makecell[t]{10\\(Dep)} & \makecell[t]{Moderate\\(Dep)}  & \makecell[t]{Severe\\(Dep)}   & Model’s judgment is appropriate. Participants’ resignation (e.g., “Isn’t everyone like this?”) may have contributed to underreporting on the PHQ-9.\\   \addlinespace[0.2pt]
  & 9   & \makecell[t]{4\\(Norm)} & \makecell[t]{19\\(Dep)} & \makecell[t]{Mild\\(Dep)}     & \makecell[t]{Moderate\\(Dep)}  & Model’s prediction is reasonable, but confidence should be moderated due to unclear current mood despite strong depressive risk factors.\\   \addlinespace[0.2pt]
  & 10 & \makecell[t]{2\\(Norm)} & \makecell[t]{10\\(Dep)} & \makecell[t]{Mild\\(Dep)}     & \makecell[t]{Moderate\\(Dep)}  & Model’s judgment is appropriate. However, distressing words are expected when referring to a deceased son, and they seem to have been overweighted in evaluating current mood. Confidence should be lower than 1. \\   \addlinespace[0.2pt]
  & 11   & \makecell[t]{1\\(Norm)}  & \makecell[t]{15\\(Dep)} & \makecell[t]{Moderate\\(Dep)}  & \makecell[t]{Severe\\(Dep)}    & Model’s judgment is appropriate. Strong depressive risk factors were noted, but unclear current mood suggests confidence should be adjusted.\\   \addlinespace[0.2pt]
  & 12 & \makecell[t]{1\\(Norm)}  & \makecell[t]{24\\(Dep)} & \makecell[t]{Moderate\\(Dep)}  & \makecell[t]{Moderate\\(Dep)}  & Model’s judgment is appropriate, as it correctly identified clear childhood adversity as a depression risk factor. However, limited information on current mood suggests confidence should be low.\\ \addlinespace[3pt] \midrule
\addlinespace[3pt]
\multirow{13}{*}{\begin{tabular}[c]{@{}l@{}}Psychiatrists\\disagreement\end{tabular}}  
  & 13 &  \makecell[t]{11\\(Dep)} & \makecell[t]{0\\(Norm)} & Normal & \makecell[t]{Mild\\(Dep)} & \textbf{Psychiatrist A}: Model’s judgment is appropriate, as depressive mood is not evident in the text. \newline
\textbf{Psychiatrist B}: Keywords related to COVID-19 isolation reflect recent stress episodes, but the model failed to capture their significance.\\   \addlinespace[0.2pt]
  & 14 & \makecell[t]{5\\(Dep)}  &  \makecell[t]{0\\(Norm)} & Normal & \makecell[t]{Mild\\(Dep)} & Model judgment is appropriate. PHQ-9 score likely due to participants’ non-mood-related factors.\\   \addlinespace[0.2pt]

  & 15 &  \makecell[t]{0\\(Norm)}  & \makecell[t]{10\\(Dep)} & Normal & \makecell[t]{Severe\\(Dep)} & \textbf{Psychiatrist A}: Model misinterprets past symptoms as current; misses signs of resilience. \newline
\textbf{Psychiatrist B}: Judgment appropriate, as participant frequently mentions panic and depression.\\   \addlinespace[0.2pt]
  & 16 &  \makecell[t]{0\\(Norm)} & \makecell[t]{15\\(Dep)} & Normal & \makecell[t]{Mild\\(Dep)} & \textbf{Psychiatrist A}: Model failed to consider participants’ personality and non-pathological coping style. Reluctance to share and solitary stress coping were misinterpreted as depressive. \newline
\textbf{Psychiatrist B}: Model’s judgment is appropriate, but overall content of loneliness suggests mild depressive mood. (Model classified as severe)\\ 

\bottomrule
\end{tabular*}
\label{tab4}
\end{sidewaystable*}

\section{Discussion}\label{sec4}
This study demonstrates the potential of leveraging LLMs for the early screening of depression.
We constructed a novel dataset comprising 3,699 retrospective narratives of happiness and distress, along with 265 entries from VEMOD.
Building on these datasets, we developed DepressLLM, a domain-adapted model that outperformed generic LLMs in depression classification.
The GPT-4-based DepressLLM consistently outperformed the GPT-3.5-based model, reflecting the benefits of recent advancements in model architecture and instruction alignment. As LLMs continue to advance, their accuracy and reliability in mental health prediction are expected to improve.

Although OpenAI models demonstrate robust predictive performance, their API-based access limits deployment flexibility and raises privacy concerns, particularly in sensitive areas such as mental health. Furthermore, We provide open-source-based DepressLLMs that demonstrate comparable classification performance. The LLaMA-3.3 based DepressLLM consistently outperformed smaller-scale models, including those based on Phi-4 14B and Qwen3-32B, reinforcing the benefits of increased model size. Further development of open‑source models is expected to enhance both predictive accuracy and functional versatility.

To enhance predictive reliability and performance, we applied the SToPS method, which aggregates token-level probabilities across candidate outputs to compute both predictions and confidence scores. By training the model on continuous PHQ‑9 scores, we guided it to capture subtle variations in depressive severity without relying on strict classification thresholds. The combination of continuous supervision with PHQ-9 scores and token-level probability summation contributed to both improved performance and greater interpretable confidence in depression classification.

Fine-tuning LLMs relies primarily on the relevance and quality of the training data \cite{touvron2023llama}. To leverage this dependency, we designed and collected a dataset based on individual personal experiences, capturing both happy and distressing memories. For effective fine-tuning for depression  screening, the dataset must contain linguistic characteristics representative of depressive symptoms. Individuals with depression tend to use more negative emotion words and fewer positive ones while describing their lives \cite{behdarvandirad2022depression,hur2024language,weintraub2023word}. To exploit this phenomenon more effectively, we  designed and collected a dataset grounded in individuals’ personal experiences, capturing emotionally rich autobiographical narratives that encompass both happy and distressing memories. These retrospective narratives likely enabled the model to learn how past experiences shaped current emotional states, which is particularly important because early life adversity and other traumatic experiences are well-established risk factors for depression \cite{li2016maltreatment}. Notably, the model demonstrates robust generalization to other datasets with different formats and contexts, such as daily mood reports (VEMOD) and structured clinical interviews (DAIC-WOZ). This robustness may be attributed to several factors: first, the model may have learned how emotional content is expressed across diverse narrative contexts \cite{tang2024decoding}; second, it may have captured individual language styles that reflect affective coloring \cite{trifu2024linguistic}; and third, training on both positive and negative memories likely enhanced its ability to interpret subtle signals of mood across a wide spectrum \cite{ren2021depression}. These findings highlight the value of a data-centric approach for building interpretable and context-aware models for mental health prediction.

An independent psychiatric review of high-confidence misclassified cases revealed that the model’s predictions were often more consistent with clinical judgment than with participants’ self-reported PHQ-9 scores. This suggests that high-confidence outputs may, in some cases, better reflect clinical reality. Of the 16 cases reviewed, both psychiatrists determined that 12 cases aligned closely with the model's interpretation. Based on the participants’ narratives, the psychiatrists identified plausible reasons for these discrepancies, many of which reflected the known limitations of self-report instruments and underscored the potential utility of language-model-based interpretations as a complementary signal in depression assessment. In the remaining four high-confidence misclassified cases, no instances in which both psychiatrists judged the model’s prediction to be incorrect existed. Instead, the clinicians disagreed with each other, highlighting the absence of an absolute ground truth, even among trained experts. However, the high confidence assigned by the model to these ambiguous cases indicates a limitation in its ability to recognize uncertainty. This observation suggests the need for further refinement of the confidence estimation mechanism, particularly in contexts where linguistic signals are subtle, context-dependent, or open to multiple interpretations. Simultaneously, the variation in expert judgments underscores the potential value of language models as supportive tools that can offer consistent and reproducible interpretations in domains where subjectivity is prevalent.

A key strength of this study lies in its integration of explainability and confidence estimation, both of which are essential for the clinical applicability of AI-based mental health tools. The model produced concise natural-language explanations that highlighted key linguistic cues, whereas the confidence scoring mechanism provided calibrated estimates of predictive certainty, enabling the down-weighting of ambiguous predictions and supporting more cautious deployment in real-world settings. This is particularly valuable in mental health contexts, where helping individuals understand their emotional states can improve insight, reduce stigma, and enhance treatment engagement. In the expert review of high-confidence misclassified cases, most model explanations were deemed clinically appropriate, reinforcing the interpretability and trustworthiness of the model’s outputs. Furthermore, the reviewers identified areas for improvement, including inadequate handling of temporal context, lack of attention to protective factors, and an absence of uncertainty expression in low-content narratives. These insights offer practical guidance for refining both the explanation and confidence mechanisms and for improving the alignment between model reasoning and clinical judgment.

However, this study has several limitations.  Retrospective narratives of happiness and distress may lack temporal sensitivity because individuals might provide similar responses within short time intervals, regardless of actual emotional fluctuations.
Moreover, the PHQ-9 scores were based on participants’ self-assessments, which may not accurately reflect their actual mental health status owing to factors such as limited emotional awareness or social desirability bias. We anticipate that training with labels more closely aligned with clinical ground truth could further improve model performance. 

\section{Methods}\label{sec5}
\subsection{Datasets}
\subsubsection{TREND-P dataset}
The TREND-P dataset was constructed by the corresponding author at the Chonnam National University Hospital (CNUH) and Chonnam National University Hwasun Hospital (CNUHH) to identify digital biomarkers of mental disorders using a transdiagnostic approach, thereby ensuring high clinical reliability and ecological validity. The dataset integrates multimodal information from individuals with psychiatric conditions and healthy controls, including psychological assessments, video-recorded audio interviews, heart rate variability, vital signs, actigraphy, blood biomarkers, and smartphone-based digital behavior data. Data collection began on August 2, 2021; for the present analysis, data collected up to January 20, 2025, were included.
Among the speech tasks in the dataset, this study focused on a free narrative task in which participants were asked to recall and describe in detail one personally meaningful joyful memory and one painful memory, each lasting at least three minutes. This task was developed based on a diverse review of the existing literature on voice analysis in depression \cite{wang2019acoustic}. When participants experienced difficulty initiating or sustaining speech, trained clinical interviewers provided minimal prompts to facilitate continued narration without influencing the content. All speeches were recorded and transcribed into text. Only the transcribed textual data were used in this analysis.
All participants provided written informed consent before participation. The study protocol was approved by the Institutional Review Boards of CNUH and CNUHH (approval numbers: CNUH-2021-243, CNUH-2022-216, CNUHH-2021-117, and CNUHH-2022-126).

\subsubsection{VEMOD dataset}
The VEMOD dataset was constructed by the corresponding author to identify digital biomarkers of mental health status through high-frequency ecological momentary assessments of a high-stress occupational group. Participants were recruited from three call centers and one public agency. Data were collected on-site at the participants’ workplaces between August 3, 2023, and January 25, 2024. All participants provided written informed consent. The study protocol was approved by the Institutional Review Boards of CNUH and CNUHH (approval numbers: CNUH-2023-156 and CNUHH-2023-118).
At baseline, the participants completed in-person assessments, including sociodemographic, psychiatric, and personality questionnaires. Subsequently, a custom-developed EMA mobile application was installed for real-time data collection. During the two-week study period, participants completed EMA tasks three times daily: morning (08:00) sessions included a 10-point mood rating and a voice description of the current emotional state; midday (13:00) and evening (18:00) sessions included responses to two stress-related questions and voice descriptions of stressful experiences. The evening session also included a repeated mood rating, two modified PHQ-2 items adapted for daily assessment, and one anger-related item. Throughout the two weeks, participants wore a Fitbit device continuously to collect step count and heart rate variability (HRV) data.
In the present analysis, we used only the transcribed textual data from the EMA voice recordings as a test set, comprising 265 participants.

\subsubsection{DAIC-WOZ dataset}
The DAIC-WOZ dataset \cite{gratch2014distress}, used as an external test set in this study, is a clinical interview corpus designed to support the diagnosis of psychological distress conditions, such as anxiety, depression, and post-traumatic stress disorder. It is provided in a multimodal format including text, video, and audio; in this work, we focused specifically on the text modality and the PHQ-8 \cite{kroenke2009phq} results for depression symptom classification. The corpus comprises 189 interviews. 

\subsection{Fine-tuning DepressLLM and Estimating Confidence with SToPS}
DepressLLM was trained to predict the PHQ-9 score, and the resulting probability distribution over tokens corresponding to the PHQ-9 range (0–27) was used to derive the binary depression classification. We fine-tuned LLMs, including OpenAI’s GPT-4 \cite{achiam2023gpt} and three open-source models (Phi-4 14B \cite{abdin2024phi}, Qwen3-32B \cite{yang2025qwen3}, and LLaMA-3.3 70B \cite{grattafiori2024llama}). For fine-tuning, we constructed an instruction-based prompt format in which the model received a system message that specified the task of predicting a PHQ-9 score, along with a user message containing a participant’s narrative.

To quantify both the model’s predicted depression probability and the confidence of each prediction, we proposed SToPS, which computes the cumulative probability across all score tokens greater than or equal to a decision cutoff \( d \). The predicted probability of depression is defined as follows:
\[
P(\text{Depression}) = \sum_{s \geq d} p(s)
\]
where \( p(s) \) denotes the model-assigned probability for score token \( s \). The SToPS-based confidence score is then calculated as follows:
\[
\text{Confidence} = 2 \cdot \left| P(\text{Depression}) - 0.5 \right|
\]
We considered the differences in tokenization across the models. In multi-digit tokenization models (e.g., Phi-4, LLaMA-3.3, and GPT-4.1), each score in the range of 0–27 is typically represented as a single token, allowing us to directly use the token-level probabilities \( p(s) \).
In contrast, Qwen3-32B uses single-digit tokenization; therefore, two-digit scores (10–27) are output as a sequence of two-digit tokens: \(d_1\) (the first digit) and \(d_2\) (the second digit). We then compute the joint probability of the score \(s = d_1d_2\) via the chain rule, as follows:
\[
p(s) = p(d_1) \cdot p(d_2 \mid d_1).
\]

\subsection{Experimental Setup and Baselines}
We fine-tuned the GPT-based models usinged OpenAI’s fine-tuning API, whereas the open-source models were fine-tuned in a local training environment equipped with an NVIDIA A100 GPU (80 GB memory). We applied low-rank adaptation (LoRA) \cite{hu2022lora} to fine-tune the Phi-4 and Qwen3 models. The LLaMA-3.3 model was fine-tuned using Quantized LoRA (QLoRA) \cite{dettmers2023qlora}, which integrates 4-bit quantization with LoRA to enable memory-efficient training of large-scale language models. 
For the OpenAI models, we relied on the platform’s default fine-tuning settings, which automatically optimized hyperparameters based on dataset size. For Phi-4, Qwen3, and LLaMA-3.3, we used a LoRA rank of 16, a learning rate of 2e-4, a per-device batch size of 2, and gradient accumulation steps of 4, resulting in an effective batch size of 8. Additionally, during zero-shot inference with LLaMA-3.3, we employed a 4-bit quantized version of the model to accommodate GPU memory constraints.
Performance comparisons were averaged over three runs (seeds 0, 1, and 2), and the zero‑shot evaluation was conducted deterministically (temperature = 0). Subsequent analyses (Sections 2.4–2.6) were performed on the seed 0 model, with depression defined by a clinical cutoff of 5.

Additionally, as an embedding-based baseline, we generated embeddings with OpenAI’s text-embedding-3-large and trained an XGBoost classifier \cite{chen2016xgboost}.  For domain-specific baselines, we used the fine-tuned MentaLLaMA-chat-13B model \cite{yang2024mentallama} for inference, whereas the MentalBERT and MentalRoBERTa models \cite{ji2021mentalbert} were further fine-tuned using the TREND-P training set.

\subsection{Class-normalized frequency of significant lexical cues}
We normalized the frequencies of the significant words and phrases that DepressLLM identified as the most informative lexical cues for its predictions. For each word/phrase \( w \), the frequency within each predicted class \( c \) was computed as a class-normalized percentage, as follows:
\[
\mathrm{Percentage}_{w,c} = \frac{n_{w,c}}{N_c} \times 100,
\qquad c \in \{\text{Normal}, \text{Depression}\},
\]
where \( n_{t,c} \) denotes the number of utterances containing word/phrase \( w \) classified as class \( c \), and \( N_c \) is the total number of utterances assigned to class \( c \).
For the TREND-P dataset, which includes both memories of happiness and distress from each participant, an additional breakdown by memory-prompt type was applied. This produced three distinct subgroups based on memory context (happy, distressed, and both), yielding six mutually exclusive combinations of predicted classes and contexts.

\bmhead{Data availability}
The datasets generated and analyzed during the current study are available from the corresponding author on reasonable request.

\bmhead{Code availability}
We provide our LoRA adapter weights on Hugging Face at https://huggingface.co/SehwanMoon/DepressLLM-llama-3.3 70B. Furthermore, we provide a web demo of DepressLLM (https://depressllm.streamlit.app/).

\bmhead{Acknowledgements}
This work was supported by Electronics and Telecommunications Research Institute (ETRI) grant funded by the Korean government (25ZK1100, Honam regional industry-based ICT convergence technology advancement support project). This study was supported by the Bio\&Medical Technology Development Program of the National Research Foundation (NRF) funded by the Korean government (MSIT) (No.RS-2024-00440371).
We sincerely thank all individuals who participated in this study. Their openness and contributions made this research possible, and we deeply respect their willingness to share their personal reflections on moments of happiness and distress.

\bmhead{Author contributions}

S.M., J.W.K., and M.J. designed and conducted this study. S.M., J.W.K., and M.J. contributed to writing the original draft. S.M. ran experiments and created figures. H.J.K., S.W.K., J.M.K., I.S.Shin, A.L., and J.E.K. contributed to validation. J.W.K. and M.J. were responsible for data acquisition and curation. All authors contributed to reviewing and editing the manuscript.

\bmhead{Competing interests}
The authors declare no competing interests.

\bibliography{sn-article}
\newpage



\section*{Supplementary material}

# Supplementary information for DepressLLM: Interpretable Domain-Adapted Language Model for Depression Detection from Real-World Narratives

Sehwan Moon[1], Aram Lee[1], Jeong Eun Kim[1], Hee-Ju Kang[2], Il-Seon Shin[2], Sung-Wan Kim[2*], Jae-Min Kim[2*], Min Jhon[2*], Ju-Wan Kim[2*]

[1]AI Convergence Research Section, Electronics and Telecommunications Research Institute, Republic of Korea.
[2]Department of Psychiatry, Chonnam National University Medical School, 160 Baekseoro, 12 Dong-gu, Gwangju, 61469, Republic of Korea.

*Corresponding author(s). E-mail(s): shalompsy@hanmail.net; jmkim@chonnam.ac.kr; minjhon@chonnam.ac.kr; tarot383@naver.com; Contributing authors: sehwanmoon@etri.re.kr;

# Contents

# List of Tables

# 1 Prompt Instructions

In this study, we designed the prompt instructions as follows.

**Instruction for Score-Based Zero-Shot Classification**

> You will be given a transcript of a participant talking about happiness and distress.
> Classify the transcript into one of the PHQ-9 scores (0–27).
> Respond with only the score as an integer. Do not include any other text.

**Instruction for Binary Classification**

> You will be given a transcript of a participant talking about happiness and distress.
> Classify the transcript into one of the two classes: 0 (normal) or 1 (depression).

**Instruction for SToPS-Based Fine-Tuning**

> You will be given a transcript of a participant talking about happiness and distress.
> Classify the transcript into one of the PHQ-9 scores (0–27).

**Instruction for SToPS-Based Fine-Tuning with Explanation**

> You will be given a transcript of a participant talking about happiness and distress.
>
> 1. Classify the transcript into one of the PHQ-9 scores (0–27).
> 2. Write a brief explanation for your prediction by referring to evidence from the transcript.
> 3. Highlight all significant words or phrases that influenced your decision, separated by commas.
>
> Example output format:
> <PHQ-9 score as integer>
> Explanation: <Brief explanation, citing specific evidence from the transcript.>
> Significant words/phrases: <phrase 1>, <phrase 2>, ...

**Instruction for SToPS-Based Fine-Tuning with Explanation and Confidence**

> You will be given a transcript of a participant talking about happiness and distress.
>
> 1. Classify the transcript into one of the PHQ-9 scores (0–27).
> 2. Write a brief explanation for your prediction by referring to evidence from the transcript.
> 3. Highlight all significant words or phrases that influenced your decision, separated by commas.
> 4. Provide a confidence score (as a percentage) indicating how certain you are of your prediction.
>
> Example output format:
> <PHQ-9 score as integer>
> Explanation: <Brief explanation, citing specific evidence from the transcript.>
> Significant words/phrases: <phrase 1>, <phrase 2>, ...
> Confidence score: <percentage>%

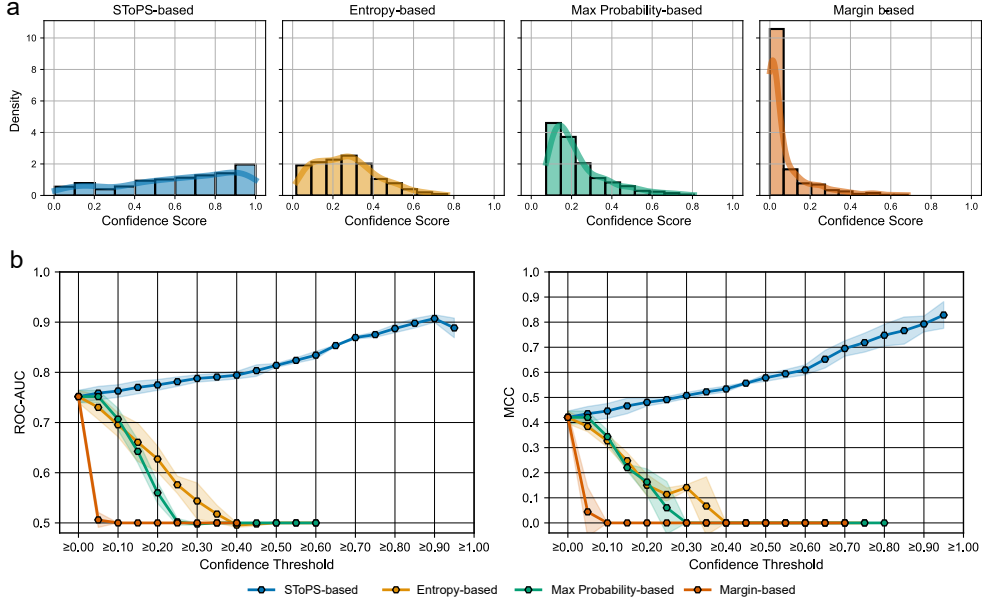

**Fig. S1 Comparison of confidence estimation methods (SToPS-based, entropy-based, max token probability–based, and margin-based).** (a) distribution of confidence scores for each method, and (b) classification performance across varying confidence thresholds.

## 2 Comparison of Confidence Estimation Methods

We compared four confidence estimation methods, including three existing approaches and the proposed SToPS method. Each method computes a confidence score based on the predicted probability distribution of the model on the PHQ-9 score tokens. In this experiment, we adopted a score-based classification method to compute the depression probability and used a clinical cutoff of PHQ-9 $\geq 5$. The classification outputs remained the same across all methods; only the computation of confidence scores differed.

The confidence method based on entropy quantifies uncertainty using the normalized Shannon entropy of the predicted distribution [1]. Confidence score is defined as:

$$\text{Entropy-based confidence} = 1 - \frac{H(p)}{\log K}, \quad H(p) = -\sum_{s=0}^{K-1} p(s) \log p(s)$$

where $K$ is the number of score tokens, and $p(s)$ denotes the probability assigned to score token $s$. Lower entropy corresponds to higher confidence. The max token probability–based confidence uses the highest probability among all predicted tokens as the confidence score:

$$\text{Max token probability–based confidence} = \max_s p(s)$$

The margin-based confidence method considers the difference between the probabilities of the two highest-probability prediction tokens:

$$\text{Margin-based confidence} = p_{\text{top-1}} - p_{\text{top-2}}$$

Figure S1 illustrates the distribution of estimated confidence scores for each method, along with the classification performance at varying confidence thresholds.

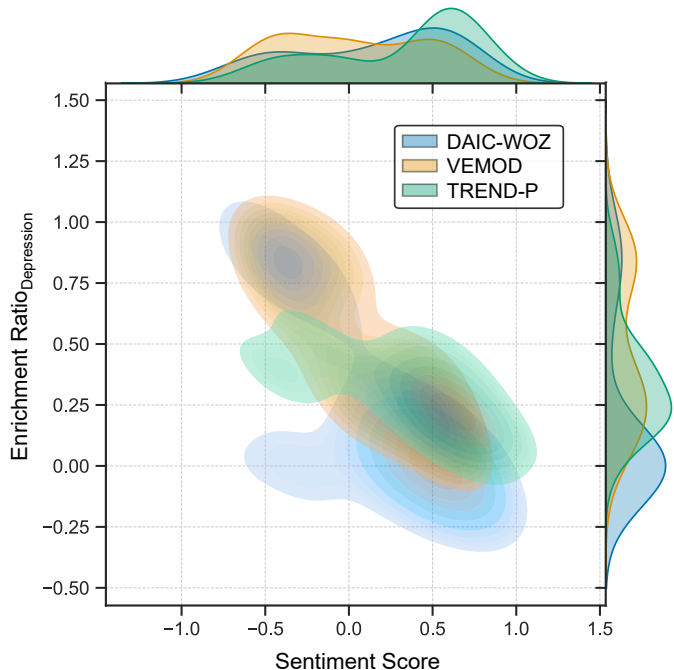

**Fig. S2 2D kernel density estimate (KDE) showing the joint distribution of sentiment polarity and the depression-class enrichment ratio for significant words/phrases across three datasets (DAIC-WOZ, VEMOD, and TREND-P).** The x-axis denotes the VADER sentiment score of each word/phrase (−1 to +1). The y-axis denotes the depression-class enrichment ratio. An enrichment ratio near 1 indicates terms significant exclusively in depression-class predictions, and a ratio of 0.5 denotes no class preference.

## 3 Analysis of Sentiment and Class-Specific Enrichment Ratios

We examined the relationship between sentiment scores and class-specific patterns by analyzing the 50 most frequently extracted words and phrases. Sentiment was computed with the Valence Aware Dictionary and sEntiment Reasoner (VADER) model [2]. Each term was assigned a sentiment score from –1 (negative) to +1 (positive). For

each word/phrase, we also computed the depression-class enrichment ratio as:

$$\text{Enrichment Ratio}_{\text{Depression}} = \frac{P_{\text{Depression}}}{P_{\text{Normal}} + P_{\text{Depression}}}$$

where $P_{\text{Normal}}$ and $P_{\text{Depression}}$ indicate the normalized frequencies of the predicted classes of words and phrases within each dataset. The ratio measures how often a significant word or phrase is used in depression predictions than in normal ones. When the enrichment ratio is near 1, the term is specific to depression-class predictions; near 0, it is specific to normal-class predictions. A value of 0.5 indicates no class preference. Figure S2 shows a 2D kernel-density plot of sentiment polarity versus enrichment ratio.

**Table S1** High-Confidence Misclassification Cases: Model Outputs and Psychiatrist Reviews

| Field | Value |
| --- | --- |
| Case # | 1 |
| Actual PHQ-9 | 5 |
| Confidence score | 0.955991198239648 |
| Predicted PHQ-9 | 0 |
| Explanation | The participant describes a life filled with happiness, joy, and gratitude, despite having experienced significant hardships and losses. She expresses contentment with her current life, enjoys her work, finds joy in daily activities, and maintains strong social connections. While she recounts past distressing events, she does not report current symptoms of depression such as persistent sadness, loss of interest, sleep or appetite disturbances, or feelings of hopelessness. Her narrative is characterized by resilience and positive coping. |
| Significant words/phrases | happiest and most joyful memory, joyful times and happy memories, so happy, enjoy life, every day...is fun and interesting, living at home is also so, so happy, being healthy enough to do these things, makes me so happy, I've let go of all that, I was happy to send her off that way, grateful, contentment |
| Psychiatrist A | Normal |
| Psychiatrist B | Normal |
| Key physician comments (condensed) | Appropriate model judgment. PHQ-9 likely reflects participants' non-mood symptoms. |

| Field | Value |
| --- | --- |
| **Field** | **Value** |
| Case # | 2 |
| Actual PHQ-9 | 6 |
| Confidence score | 0.9568000000000001 |
| Predicted PHQ-9 | 0 |
| Explanation | The participant expresses consistent happiness, gratitude, and satisfaction with life, family, and health. There is no mention of persistent sadness, loss of interest, or other depressive symptoms. The brief mention of distress is minimized and not described as ongoing or significant. |
| Significant words/phrases | always happy and joyful, everything in life was truly great, thankful, joyful, enjoyable, life has been truly fun and happy, hope everyone stays healthy, always live happily and well, It was so hard, and it felt so distressing, but I don't think there were that many difficult things |
| Psychiatrist A | Normal |
| Psychiatrist B | Normal |
| Key physician comments (condensed) | Model's judgment is appropriate based on interview content alone, but the limited mention of distress appears to make accurate assessment difficult. |

| Field | Value |
| --- | --- |
| Case # | 3 |
| Actual PHQ-9 | 4 |
| Confidence score | 0.9821553884711781 |
| Predicted PHQ-9 | 10 |
| Explanation | The participant describes significant distress and symptoms consistent with moderate depression. They report persistent sadness, lack of satisfaction, constant pain, forgetfulness, and a history of suicidal ideation and attempts. They also mention feeling that everything has become unpleasant, and express ongoing anxiety and worry about their husband. However, they also describe some positive aspects, such as pride in their children and support from their daughter, which may mitigate the severity of their depression. |
| Significant words/phrases | forget things all the time, my mind is scattered, I feel like I haven't really lived my life, it makes me sad, everything seems to have become unpleasant, my body is in constant pain, there's no part of me that doesn't hurt, the saddest thing, life revolves around alcohol, no real sense of satisfaction, I was hit a lot, I didn't want to live anymore, tried to end it, constantly worried and anxious |
| Psychiatrist A | Moderate (Dep) |
| Psychiatrist B | Severe (Dep) |
| Key physician comments (condensed) | Model judgment is appropriate. Direct mention of sadness and depression suggests possible misreporting on PHQ-9 items. |

| Field | Value |
| --- | --- |
| Case # | 4 |
| Actual PHQ-9 | 4 |
| Confidence score | 0.9917678947896797 |
| Predicted PHQ-9 | 10 |
| Explanation | The participant describes ongoing symptoms of depression, including persistent sadness, anxiety, and difficulty recovering from the loss of a child, despite medication. They mention that living is "so difficult," and that they "can't seem to recover from this illness," indicating moderate depressive symptoms. However, they also mention some positive experiences, such as finding joy in their grandchildren and trying to take care of their other children, suggesting that their symptoms are not in the severe range. |
| Significant words/phrases | depression, can't seem to recover, pain, heart aches, sense of anxiety, uneasiness, so hard for me, living like this is so difficult, can't forget, keep taking the medication |
| Psychiatrist A | Mild (Dep) |
| Psychiatrist B | Moderate (Dep) |
| Key physician comments (condensed) | Appropriate model judgment, but past events are overemphasized. Low PHQ9 likely due to participants' poor mood awareness. |

| Field | Value |
| --- | --- |
| Case # | 5 |
| Actual PHQ-9 | 1 |
| Confidence score | 1.0 |
| Predicted PHQ-9 | 13 |
| Explanation | The participant describes significant distressing experiences, including childhood trauma, domestic violence, divorce, and ongoing concerns about their son. They also report an absence of happy memories and state, "I don't think there was anything particularly worth considering as happy," suggesting persistent low mood and anhedonia. These factors are consistent with moderate depression. |
| Significant words/phrases | I don't think I have such memories, I don't think there was anything particularly worth considering as happy, my mom was hit by my dad almost every day, I would hug my mom with my body so that my dad couldn't hit her, those memories of trying to protect her are the most painful, the years before my divorce were very difficult, my son made a wrong choice and went down a certain path, he's still in that situation, those memories of struggling with that are very painful |
| Psychiatrist A | Mild (Dep) |
| Psychiatrist B | Mild (Dep) |
| Key physician comments (condensed) | Content is insufficient to assess mood, but model judgment is appropriate given childhood adversity. The model should have noted the lack of information. |

| Field | Value |
| --- | --- |
| Case # | 6 |
| Actual PHQ-9 | 4 |
| Confidence score | 0.9693601682186843 |
| Predicted PHQ-9 | 8 |
| Explanation | The participant describes significant distress related to her husband's financial irresponsibility, betrayal, and ongoing family obligations, leading to insomnia and persistent stress. She reports spending several months with negative thoughts about her relationship and feeling overwhelmed by her responsibilities. However, she also describes positive experiences, such as happiness with her home and her daughter's achievements, suggesting that while her symptoms are moderate, they are not at the most severe level. |
| Significant words/phrases | insomnia, under extreme stress, couldn't stand the sight of my husband, hated him so much, spent several months thinking about that, makes me so sad, suffer from insomnia, stress is my husband and his family |
| Psychiatrist A | Mild (Dep) |
| Psychiatrist B | Mild (Dep) |
| Key physician comments (condensed) | Model judgment is appropriate. Reason for participants' low PHQ-9 score is unclear. |

| Field | Value |
| --- | --- |
| Case # | 7 |
| Actual PHQ-9 | 1 |
| Confidence score | 0.9843404938767315 |
| Predicted PHQ-9 | 10 |
| Explanation | The participant describes significant distress, including a period of severe physical symptoms, weight loss, and inability to eat or perform daily activities, which are consistent with moderate depression. They mention a diagnosis of somatization disorder and depression, and although they are now improving and able to manage some daily activities, they still find things "a bit difficult." The presence of ongoing symptoms and the need for treatment suggest a moderate level of depressive symptoms. |
| Significant words/phrases | couldn't eat or do anything, lost more than 10% of my body weight, so exhausted, couldn't breathe, all the strength left my hands and feet, it was so hard, still a bit difficult, receiving treatment, improving a lot now |
| Psychiatrist A | Mild (Dep) |
| Psychiatrist B | Moderate (Dep) |
| Key physician comments (condensed) | Model's judgment is appropriate, but key words should account for temporal context. Low PHQ-9 may reflect somatization. |

| Field | Value |
| --- | --- |
| **Field** | **Value** |
| Case # | 8 |
| Actual PHQ-9 | 2 |
| Confidence score | 0.9599158232287803 |
| Predicted PHQ-9 | 10 |
| Explanation | The participant describes significant distress, including feeling like wanting to die, crying a lot, and feeling alone due to the loss of family members. They also mention a lack of motivation, lying down at home a lot, and being annoyed or stressed by family matters. However, they still engage in some activities like shopping, hanging out with friends, and housework, and experience periods of happiness, especially in the past. The presence of both moderate depressive symptoms and some functional activity suggests a moderate level of depression. |
| Significant words/phrases | my body doesn't keep up anymore, my body is still in a pretty bad state, I just lie down at home a lot, I get annoyed, I don't feel like moving, I wanted to just leave everything behind, it's truly painful, I even feel like I want to die, I cry a lot when I'm alone, all my parents and siblings are gone, I'm left alone, I feel like leaving the worldly life behind, those thoughts repeat again and again |
| Psychiatrist A | Moderate (Dep) |
| Psychiatrist B | Severe (Dep) |
| Key physician comments (condensed) | Model's judgment is appropriate. Participants' resignation (e.g., "Isn't everyone like this?") may have contributed to underreporting on the PHQ-9. |

| Field | Value |
| --- | --- |
| Case # | 9 |
| Actual PHQ-9 | 4 |
| Confidence score | 1.0 |
| Predicted PHQ-9 | 19 |
| Explanation | The participant describes a history of severe depression, including a period of reclusive behavior ("hikikomori"), intense self-hatred, and significant functional impairment (not eating for days, isolating from family). These symptoms suggest persistent and severe depressive symptoms, likely including anhedonia, depressed mood, sleep/appetite disturbance, low energy, feelings of worthlessness, and social withdrawal. The narrative indicates ongoing distress and impairment, consistent with a high PHQ-9 score. |
| Significant words/phrases | severe obsessive-compulsive disorder and depression worsened, lived a reclusive life, hikikomori, intense self-hatred, I don't even deserve to eat, go days without eating, barely managing to eat one meal after starving for several days, lock my door and stay inside, mom knocking on my door, crying for an hour straight |
| Psychiatrist A | Mild (Dep) |
| Psychiatrist B | Moderate (Dep) |
| Key physician comments (condensed) | Model's prediction is reasonable, but confidence should be moderated due to unclear current mood despite strong depressive risk factors. |

| Field | Value |
| --- | --- |
| Case # | 10 |
| Actual PHQ-9 | 2 |
| Confidence score | 0.9560306575231948 |
| Predicted PHQ-9 | 10 |
| Explanation | The participant describes ongoing, significant distress related to the loss of her son, which continues to affect her deeply even after fifteen years. She expresses persistent sadness, difficulty coping, and a sense of emptiness. While she mentions some happy memories from the past, her current emotional state is dominated by grief and pain, which likely impacts her daily functioning and mood. |
| Significant words/phrases | most difficult and painful thing, sending our son to heaven, most agonizing and hardest thing, my chest was completely pierced through, I can't bear it, I absolutely can't do that, my heart isn't like that, I'm in pain, no way to express it in words |
| Psychiatrist A | Mild (Dep) |
| Psychiatrist B | Moderate (Dep) |
| Key physician comments (condensed) | Model's judgment is appropriate. However, distressing words are expected when referring to a deceased son, and they seem to have been overweighted in evaluating current mood. Confidence should be lower than 1. |

| Field | Value |
| --- | --- |
| Case # | 11 |
| Actual PHQ-9 | 1 |
| Confidence score | 1.0 |
| Predicted PHQ-9 | 15 |
| Explanation | The participant describes significant distress, including multiple suicide attempts, hospitalization, persistent negative thoughts, and physical self-harm. However, they also report many positive experiences and sources of happiness, such as travel, family, friends, and enjoyable activities. This suggests moderate to moderately severe depression, but not the most severe category, as there are still periods of enjoyment and engagement. |
| Significant words/phrases | kept attempting suicide every day, wrists were constantly scarred, surgery on my wrist was also so painful, people around me acted like it was nothing, so distressing, frustrating, painful, memories of him drinking and cursing, made harsh remarks, never received any comfort, so hard while I was hospitalized, medication didn't work, but also: happiest, enjoyable, fun, playing with the cat, cooking while playing music, traveling with family, studying with aunt, eating desserts, chatting, playing with my parents |
| Psychiatrist A | Moderate (Dep) |
| Psychiatrist B | Severe (Dep) |
| Key physician comments (condensed) | Model's judgment is appropriate. Strong depressive risk factors were noted, but unclear current mood suggests confidence should be adjusted. |

| Field | Value |
| --- | --- |
| **Case #** | 12 |
| Actual PHQ-9 | 1 |
| Confidence score | 1.0 |
| Predicted PHQ-9 | 24 |
| Explanation | The participant describes multiple severe depressive symptoms, including suicidal ideation ("That was the time I wanted to die the most", "It was so hard that I wanted to die", "I originally attempted suicide for about two days"), feelings of worthlessness ("Am I worth nothing except for money?", "What am I?"), hopelessness ("Life felt so hopeless"), and significant distress from childhood trauma and ongoing interpersonal difficulties. While the participant recalls some happy memories, the distress transcript is dominated by severe and persistent symptoms consistent with a high PHQ-9 score. |
| Significant words/phrases | "wanted to die the most", "I wanted to die", "I originally attempted suicide", "Life felt so hopeless", "felt like I was abandoned", "worth nothing except for money", "harder than being hit", "receiving too much attention from others was really hard", "studying was hard", "I really hated studying", "hopeless", "suicide attempt" |
| Psychiatrist A | Moderate (Dep) |
| Psychiatrist B | Moderate (Dep) |
| Key physician comments (condensed) | Model's judgment is appropriate, as it correctly identified clear childhood adversity as a depression risk factor. However, limited information on current mood suggests confidence should be low. |

| Field | Value |
| --- | --- |
| Case # | 13 |
| Actual PHQ-9 | 11 |
| Confidence score | 0.9625962596259626 |
| Predicted PHQ-9 | 0 |
| Explanation | The participant describes multiple positive experiences, such as promotions, buying a home without debt, and enjoying golf. The distressing experiences mentioned (wife's illness, COVID-19 test, vaccine side effects, and a past financial loss) are described as temporary and not severely impacting their overall mood or functioning. There is no evidence of persistent depressive symptoms. |
| Significant words/phrases | joyful day, happy every time I got promoted, happy, proud and happy, not exactly distressing, physically very exhausting, quite worried, fortunately, result came back negative, struggled for two days, blamed myself, felt a bit distressed |
| Psychiatrist A | Normal |
| Psychiatrist B | Mild (Dep) |
| Key physician comments (condensed) | Psychiatrist A: Model's judgment is appropriate, as depressive mood is not evident in the text. Psychiatrist B: Keywords related to COVID-19 isolation reflect recent stress episodes, but the model failed to capture their significance. |

| Field | Value |
| --- | --- |
| **Case #** | 14 |
| Actual PHQ-9 | 5 |
| Confidence score | 0.9678 |
| Predicted PHQ-9 | 0 |
| Explanation | The participant describes a generally positive outlook on life, with strong family relationships, enjoyment of daily activities, and effective coping strategies for distress. They mention experiencing distress related to their father's passing and past relationship, but also emphasize resolution, understanding, and not dwelling on negative memories. There is no evidence of persistent depressive symptoms such as loss of interest, low mood, or functional impairment. |
| Significant words/phrases | happiness in my life, happy memories, good relationship with my sons and my wife, happiest when traveling with my family, passing [certification exam]... happiest and most enjoyable moment, try to preserve them as memories, try hard to overcome, don't think those parts were something I considered extremely difficult, everything has been resolved and understood, try not to feel pain or distress, try to forget those things quickly, relieved a lot of stress through exercise |
| Psychiatrist A | Normal |
| Psychiatrist B | Mild (Dep) |
| Key physician comments (condensed) | Model judgment is appropriate. PHQ-9 score likely due to participants' nonmood-related factors. |

| Field | Value |
| --- | --- |
| Case # | 15 |
| Actual PHQ-9 | 0 |
| Confidence score | 0.9674208144796378 |
| Predicted PHQ-9 | 10 |
| Explanation | The participant describes significant distress, including symptoms of depression such as hopelessness ("I had no thoughts, no hope"), suicidal ideation ("If I just fall from here, I think I'll feel at peace"), and ongoing anxiety and panic attacks ("trouble breathing", "panic disorder", "sometimes I would feel like I couldn't breathe"). They also mention functional impairment, such as stopping driving and being unable to travel. However, they also report some positive experiences and coping mechanisms, such as faith, family support, and counseling, which may mitigate the severity of their symptoms. Overall, the symptoms described are consistent with moderate depression. |
| Significant words/phrases | no thoughts, no hope, If I just fall from here, I think I'll feel at peace, panic disorder, trouble breathing, couldn't breathe, stopped driving, haven't fully recovered, comes back again, medication, counseling |
| Psychiatrist A | Normal |
| Psychiatrist B | Severe (Dep) |
| Key physician comments (condensed) | Psychiatrist A: Model misinterprets past symptoms as current; misses signs of resilience. Psychiatrist B: Judgment appropriate, as participant frequently mentions panic and depression. |

| Field | Value |
| --- | --- |
| Case # | 16 |
| Actual PHQ-9 | 0 |
| Confidence score | 1.0 |
| Predicted PHQ-9 | 15 |
| Explanation | The participant describes a lack of interest or pleasure in activities they used to enjoy, such as traveling and interacting with nature. They also mention significant social withdrawal, lack of communication, and feelings of emptiness and isolation. The distress transcript highlights avoidance of painful memories, persistent sadness, and a desire to avoid social interaction, all of which are consistent with moderate to moderately severe depression. |
| Significant words/phrases | I don't have any thoughts, there's nothing to talk about, nothing good, nothing joyful, I don't want to talk about anything, it's painful, I don't want to meet anyone or talk to anyone, wander around alone, come back alone |
| Psychiatrist A | Normal |
| Psychiatrist B | Mild (Dep) |
| Key physician comments (condensed) | Psychiatrist A: Model failed to consider participants' personality and nonpathological coping style. Reluctance to share and solitary stress coping were misinterpreted as depressive. Psychiatrist B: Model's judgment is appropriate, but overall content of loneliness suggests mild depressive mood. (Model classified as severe) |


\end{document}